\documentclass[10pt,twocolumn,letterpaper]{article}

\usepackage{wacv}

\usepackage{graphicx}
\usepackage{amsmath}
\usepackage{booktabs}
\usepackage{multirow}
\usepackage{microtype}
\usepackage{xcolor}
\usepackage{marvosym}

\definecolor{wacvblue}{rgb}{0.21,0.49,0.74}
\usepackage[pagebackref,breaklinks,colorlinks,allcolors=wacvblue]{hyperref}
\hypersetup{pdftitle={Know When to Zoom: How Small Is Too Small for Vision-Language Models?},
  pdfauthor={Siddharth Vohra, Min Xu}}
\usepackage{xurl}

\def\wacvPaperID{*****}
\def\confName{SENSE}
\def\confYear{2027}

\title{Know When to Zoom: How Small Is Too Small for Vision-Language Models?}

\author{Siddharth Vohra\textsuperscript{1,2} \qquad Min Xu\textsuperscript{1,\Letter}\\
\textsuperscript{1}Carnegie Mellon University, Pittsburgh, PA, USA\\
\textsuperscript{2}Amazon Web Services AI Native, Pittsburgh, PA, USA\\
{\tt\small \{svohra, mxu1\}@andrew.cmu.edu}
}

\makeatletter
\def\ps@ieeenotice{\let\@oddhead\@empty\let\@evenhead\@empty\def\@oddfoot{\raisebox{-20pt}[0pt][0pt]{\parbox[t]{\textwidth}{\scriptsize \copyright\ 2027 IEEE. Personal use of this material is permitted. Permission from IEEE must be obtained for all other uses, in any current or future media, including reprinting/republishing this material for advertising or promotional purposes, creating new collective works, for resale or redistribution to servers or lists, or reuse of any copyrighted component of this work in other works.}}}\let\@evenfoot\@oddfoot}
\makeatother

\begin{document}
\maketitle
\begin{NoHyper}\let\thefootnote\relax\footnotetext{Work done by Siddharth Vohra does not relate to the position he currently holds at Amazon Web Services AI Native.\quad\textsuperscript{\Letter}Corresponding author.}\footnotetext{This study used only public benchmarks and images we generated.}\end{NoHyper}

\begin{abstract}
Online vision-language models often miss small details in large images, because the service shrinks the image, or rejects it, before the model ever sees it.
A car 20 pixels wide in an 8000 by 6000 drone photo reaches Claude about 6 pixels wide, and OpenAI's default rejects the photo outright.
We measure how much detail survives for seven models from four vendors, using an eye exam of printed codes and 601 questions about small objects in real photos.
The smallest text a model reads half the time grows from 6 pixels or less on a small image to a predicted 10 to 13 pixels on a 7680 by 4320 photo at default settings.
Zooming wins much of the detail back.
\textbf{Splitting each photo into up to nine tiles raises accuracy by 11.1 to 14.5 percentage points for Claude and Gemini}, and by avoiding rejections it lifts GPT-6.1 Sol from 50.7\% to 83.7\%.
A two-step zoom, in which the model first names the region it needs, matches or beats tiles on Gemini and comes within 2.5 points of them on Claude, with about 40\% of their image tokens.
When a detail is lost, Claude and Gemini both say they cannot tell about half the time, but they fail differently in the rest: \textbf{Gemini gives a confident wrong answer in 43\% to 48\% of these cases, Claude in only 5\% to 9\%.}
A check based on detail size, run before the call, misses most of the gain from tiles on Claude and Gemini, but Claude's own reply is a reliable trigger: tiling only after an unsure reply comes within 1.4 points of always tiling with 62\% to 70\% of the image tokens.
We turn these findings into five rules for when an agent should zoom, and release the eye exam, every model answer and code at \url{https://siddvoh.com/know-when-to-zoom}.
\end{abstract}
\thispagestyle{ieeenotice}

\section{Introduction}
\label{sec:intro}

Many questions about images turn on small details.
A satellite scene may show a vehicle only a few pixels wide, and a security camera frame may hold a license plate in one corner.
Imagine asking a vision-language model about the color of one car in an 8000 by 6000 drone photo of a parking lot, where the car is 20 pixels wide.
Claude shrinks the photo to less than a third of its width before the model sees it, so the car arrives about 6 pixels wide.
OpenAI, at its default setting, rejects the photo outright.
A user who does not know these rules cannot tell whether a wrong answer came from the model or from the shrinking.

This happens because every service we tested fits images to its own limits.
Newer Claude models shrink a large image to at most 2576 pixels on its long side~\cite{anthropic2026vision}, and Gemini turns any image into at most 280 to 2,240 tokens, the small units a model reads, depending on a setting~\cite{google2026mediaresolution}.
OpenAI's guide says that for GPT-6 Astra the default keeps the full image but rejects very large ones~\cite{openai2026imagesvision}, and the two OpenAI models we tested behave the same way.

An agent that answers such questions should therefore first ask whether the detail will survive the trip to the model.
If it will not, the agent should zoom in, by cropping or tiling the image, before it sends the question.
The hard part is knowing when.
This paper asks the three questions an agent built on these services must answer to make that decision.
How small can a detail be before a service erases it?
What does the model say when the detail is gone?
And can a simple check before the call say when to zoom?

We tested seven models: Claude Haiku 5.5 and Sonnet 5.5, Gemini 3.5 Flash-Lite and 3.8 Flash, GPT-6 Luna and GPT-6.1 Sol, and the open model Qwen3.8-27B.
From here on we call them Haiku, Sonnet, Flash-Lite, Flash, Luna, Sol and Qwen.
For the first question, we gave them an eye exam, much like the letter chart at an optometrist's office.
Each test image shows a random five-character code, printed at one of ten text sizes on a plain gray background, the \emph{canvas}.
We call the smallest text size a model reads half the time its \emph{readable size}.
We print the same text sizes on four canvases, from small to large.
The larger the canvas, the more a service may shrink it, so comparing the canvases shows how much detail the shrinking removes.
For the other two questions, we asked 601 questions about small objects in large photos, taken from three public benchmarks~\cite{wu2024vstar,wang2025divide,zhang2025mmerealworld}.
We sent each question in up to five ways: as is, at the highest setting, as 4 tiles, as 9 tiles, or as a \emph{two-step zoom}, in which the model first names the region it needs and then sees that region, cut from the original, next to an overview of the photo.

We make three contributions.
\begin{itemize}
  \item We give the readable text size for every model and setting we tested, along with the rules each service applies (\cref{tab:rules,tab:readable}, \cref{fig:eightk}).
  For Claude, the documented resize explains much of what the model can read.
  It fits Haiku closely and Sonnet less well.
  Gemini sees more than its token count.
  \item On real photos, up to nine tiles raise accuracy by 11.1 to 14.5 percentage points for Claude and Gemini, and the two-step zoom matches or beats them on Gemini with about 40\% of their image tokens (\cref{fig:accuracy}).
  When a detail is lost, both say they cannot tell about half the time, but Gemini answers the rest with high confidence (\cref{fig:lost}).
  \item We show which signals tell an agent when to zoom.
  A check on detail size, made before the call, misses most of the gain from tiles on Claude and Gemini, while on Claude, tiling only after an unsure reply nearly matches tiling every photo.
  We turn these results into five rules, and release the eye exam, every model answer and the code.
\end{itemize}

\section{What the services do to images}
\label{sec:rules}

\begin{table*}[t]
  \centering
  \footnotesize
  \setlength{\tabcolsep}{4pt}
  \begin{tabular}{@{}lllll@{}}
    \toprule
    Service and models & Setting & Rule & 12 MP photo, 4000 $\times$ 3000 & 8K photo, 7680 $\times$ 4320 \\
    \midrule
    \multirow{2}{*}{\shortstack[l]{Claude\\{\footnotesize Haiku 5.5, Sonnet 5.5}}}
      & native & long side $\le$ 2576 px, $\le$ 4,784 tokens & 2212 $\times$ 1659 & 2576 $\times$ 1449 \\
      & standard & long side $\le$ 1568 px, $\le$ 1,568 tokens & 1270 $\times$ 952 & 1456 $\times$ 819 \\
    \midrule
    \multirow{4}{*}{\shortstack[l]{Gemini\\{\footnotesize 3.5 Flash-Lite, 3.8 Flash}}}
      & low & at most 280 tokens & 266 tokens, about 1,300 px & 264 tokens, about 1,300 px \\
      & medium & at most 560 tokens & 540 tokens, about 1,700 px & 527 tokens, about 1,700 px \\
      & high or unset & at most 1,120 tokens & 1,064 tokens, about 2,300 px & 1,100 tokens, about 2,600 px \\
      & ultra high & at most 2,240 tokens & 2,160 tokens, about 3,400 px & 2,205 tokens, about 3,100 px \\
    \midrule
    \multirow{2}{*}{\shortstack[l]{OpenAI\\{\footnotesize GPT-6 Luna, GPT-6.1 Sol}}}
      & auto (default) & full size; rejected above 30,000 patches & 4000 $\times$ 3000 & rejected \\
      & high & at most 2,500 patches & 1824 $\times$ 1368 & 2105 $\times$ 1184 \\
    \midrule
    \shortstack[l]{Qwen\\{\footnotesize Qwen3.8-27B}} & host default & at most 16,384 tokens & 4000 $\times$ 3008 & 5440 $\times$ 3072 \\
    \bottomrule
  \end{tabular}
  \caption{\textbf{What each service gives the model.}
  The last two columns give the image the model receives.
  Gemini reports only tokens, so we add the long side Flash seems to keep (\cref{sec:gemini}).}
  \label{tab:rules}
\end{table*}

Before asking what models can read, we need to know what each service does to an image (\cref{tab:rules}).
Every service measures an image in tokens.
For most, a token is a small square patch of pixels, so a token limit is also a pixel limit.

\paragraph{Claude.}
Claude 4.7 and later models shrink an image until its long side is at most 2576 pixels and it fits in 4,784 tokens, each a square of 28 by 28 pixels~\cite{anthropic2026vision}.
We call this the \emph{native} setting.
Earlier Claude models used a smaller \emph{standard} limit of 1568 pixels and 1,568 tokens.
Instead of shrinking, the API can also be asked to reject an oversized image with an error~\cite{anthropic2026vision}.
To see what the native setting gains, we also test Claude at the standard setting, applying its rule ourselves.

\paragraph{Gemini.}
Gemini works differently.
Its \emph{media resolution} setting has four levels, low, medium, high and ultra high, which cap an image at 280, 560, 1,120 and 2,240 tokens~\cite{google2026mediaresolution}.
Leaving the setting unset, the default, gives the same 1,120 tokens as high.
In our measurements, the tokens form a grid shaped like the image: at high, a 4:3 image becomes 38 by 28 tokens, whether it is small or huge.
Google does not say how many pixels each token covers, so we estimate it in \cref{sec:gemini}.
In \cref{tab:rules}, the 12 MP photo uses the long side we estimate on our 4096 by 3072 canvas, and the 8K photo the one on our 8000 by 6000 canvas.

\paragraph{OpenAI.}
OpenAI's default does not shrink a large image, but it may reject it.
Its guide describes patches of 32~by~32 pixels, and says an image is rejected if it still needs more than 30,000 patches after any resizing~\cite{openai2026imagesvision}.
The guide does not cover Luna or Sol, so we measured their rules ourselves.
At the default \emph{auto} setting, both keep the image at full size.
The service rejected exactly the images that need more than 30,000 patches.
A 7680 by 4320 photo needs 32,400, so the default rejects it.
A rejected image returns an error message, and the model never answers.
The \emph{high} setting instead shrinks any larger image to at most 2,500 patches.
Both settings count 1.2 tokens per patch.

\paragraph{Qwen.}
For the open model Qwen, the service that hosts it follows the model's released image processor~\cite{qwen2026qwen38_27b}, which shrinks an image only when it needs more than 16,384 tokens, each 32 by 32 pixels.
So Qwen sees far more pixels than Claude, Gemini, or OpenAI's high setting.

We checked all these rules against the token counts the services reported.
After allowing for the tokens of our fixed text prompt, the counts these rules predict matched the reported ones for every eye-exam image the services accepted.
Gemini's image tokens also matched in all 6,447 photo calls that got an answer.
An agent can therefore estimate, before the call, how many pixels a detail will keep.
What it still needs to know is how many pixels the model needs, and that is what the eye exam measures.

\section{Study design}
\label{sec:design}

\begin{figure}[t]
  \centering
  \includegraphics[width=\linewidth]{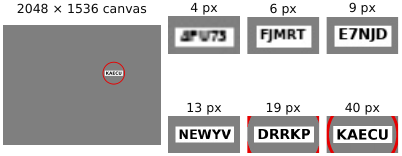}
  \caption{\textbf{The eye exam.}
  Left: a test image, with a red ring around the label.
  Right: labels at six of the ten text sizes, scaled to the same width.}
  \label{fig:example}
\end{figure}

\paragraph{Models.}
We compared the vendors at two matched tiers.
The small tier has Haiku~\cite{anthropic2026haiku55}, Flash-Lite~\cite{google2026gemini35flashlite} and Luna~\cite{openai2026gpt6astra}.
The mid tier has Sonnet~\cite{anthropic2026sonnet55}, Flash~\cite{google2026gemini38flash} and Sol~\cite{openai2026gpt61sol}.
The open model Qwen serves as a reference~\cite{qwen2026qwen38_27b}.
Every model got the same prompt, the same image files and the same answer format.
Each ran at a low reasoning level with an output limit of 8,192 tokens.
We asked once more when we could not read a reply.
Our release holds every prompt and setting, and the model version each service reported.

\paragraph{Eye exam.}
Printed codes make a clean test, because a read is either exactly right or not.
Each test image shows a random five-character code in black on a small white label (\cref{fig:example}).
The characters come from 22 letters and digits, leaving out look-alikes such as O and 0.
A guess is right only one time in 22 per character, so a lucky full read is very rare.
A thick red ring around the label stays visible after any shrinking, so the hard part is reading it, not finding it.
We measure text size as the \emph{cap height}, the height of a capital letter in pixels, at ten sizes from 4 to 128.
The canvases are plain gray, at 1024 by 768, 2048 by 1536, 4096 by 3072 and 8000 by 6000 pixels, and we name each by its long side.
Each canvas and size has 20 codes, so each setting sees 800 images.
Flash-Lite took the exam only at high and ultra high.
The model must return the five characters exactly, say whether it could read them, and give a confidence from 0 to 100.
This confidence is the number the model states, not a calibrated probability.
No model refused the task.
We sent a fixed 10\% of the images a second time, and 84\% to 100\% of the repeated reads gave the same code.

\paragraph{Real photos.}
Real questions ask about objects in whole scenes, not about a marked label.
We took 601 multiple-choice questions about small details in large photos, from three benchmarks.
V*Bench gives 191 questions about the attributes and positions of small objects in everyday scenes~\cite{wu2024vstar}.
From HR-Bench 8K we drew 150 questions, 75 from each of its two tasks~\cite{wang2025divide}.
From the Lite version of MME-RealWorld we drew 130 remote sensing and 130 monitoring questions~\cite{zhang2025mmerealworld}.
The HR-Bench and MME-RealWorld questions were drawn at random, with a fixed seed.
To every question we added one option: ``The image does not show enough detail to tell.''
Choosing it counts as wrong, and so does a rejected photo.
A model that guesses instead can therefore still be right by luck.

We sent each question in up to five ways.
\emph{As is} sends the original file at the default setting.
The 89 files over 20 MB were first re-encoded as JPEG at quality 95.
\emph{Highest setting} sends it at Gemini's ultra high level, since the other services already give their maximum.
\emph{Tiles} send an overview of the photo, shrunk to 2576 pixels on the long side if larger, followed by a 2 by 2 or 3 by 3 grid of tiles cut from the original.
Each tile overlaps its neighbors by a tenth, and the 3 by 3 grid is used only when the long side is at least 4000 pixels.
\emph{Two-step zoom} uses two calls.
The first asks the model for the region it needs, as its left, top, right and bottom edges, each a fraction of the image's width or height.
The second sends the overview together with that region, cut from the original.
The cut adds 20\% on each side and spans at least 512 pixels and a tenth of the long side.
Tiles and crops are JPEG files at quality 95, lowered to 85 to 94 for the 249 over 5 MB, which brought all but 49 crops under that size.
If the reply holds no such box, even after we ask once more, the second call sends the whole photo instead.
Like Zoomer~\cite{qian2025zoomer}, this needs no access to the model's internals.
Qwen received the photos as is only.

\paragraph{The zoom check.}
Finally, we turn the eye exam into a rule an agent could use.
The zoom check runs before each call and does not call the model.
It takes three inputs: the image size, the service's rule from \cref{tab:rules}, and the model's readable size in tokens from the eye exam.
From these it asks whether a detail of a given size, here 16 pixels tall, will still be readable once the service is done with the image.
If so, the check sends the image as is.
If not, it tries the highest setting, then 2 by 2 tiles, then 3 by 3 tiles, and picks the first under which the detail would be readable.
If none works, it uses the two-step zoom.
We use 16 pixels because it is the height of small but normal text.
We score the check on each question by the answer from the way it picked.

\paragraph{Statistics.}
For each model and setting, we fit an S-shaped curve, a logistic regression with a small ridge penalty of 0.01 on its slope, to the share of codes read exactly against the logarithm of text size.
The readable size is where this curve crosses 50\%.
We test four hypotheses, H1 to H4, each stated with its result below, and correct the tests of H2 to H4 together for multiple testing~\cite{holm1979simple}.
Intervals are 95\% intervals from 2,000 bootstrap resamples of items~\cite{efron1993introduction}, except for the rates in H2, which use exact binomial intervals~\cite{clopper1934use}.
Differences between two ways of sending the same questions use paired bootstrap intervals, and we quote them from unrounded values, so they can differ slightly from the tables.

\section{The eye exam}
\label{sec:eye}

\begin{figure*}[t]
  \centering
  \includegraphics[width=\linewidth]{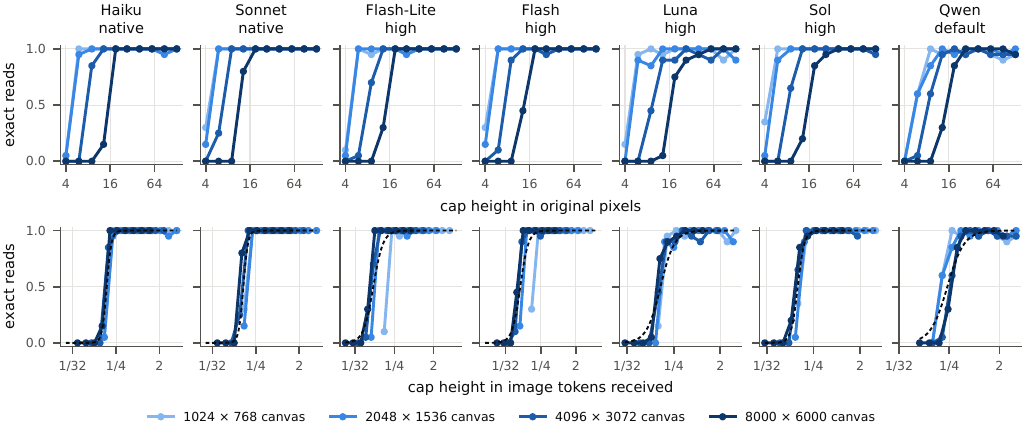}
  \caption{\textbf{Readable text size grows as the service shrinks the image.}
  Share of codes read exactly against cap height in original pixels (top) and received tokens (bottom), one line per canvas.
  Below, a dashed curve fits all four, and the lines move much closer for Claude and OpenAI.}
  \label{fig:eye}
\end{figure*}

The eye exam answers our first question: how small can a detail be before a service erases it?

\subsection{Shrinking raises the readable size}
\label{sec:shrink}

The first result is simple: the more a service shrinks an image, the larger the text must be before the model can read it.
On the smallest canvas, every model's readable size was 6 pixels or less.
Most settings shrink the two largest canvases, and there the readable size grew (\cref{fig:eye}, top).
Sonnet, for example, needs 4.2 pixels on the 1024 canvas but 12.0 pixels on the 8000 canvas.
Flash at high, which matches its default, needs 4.2 and 13.2 pixels on the same two canvases.
\Cref{tab:readable} gives the readable size for every model, setting and canvas.
At OpenAI's auto setting, the models also misread some large text, so for three entries in \cref{tab:readable} an S-shaped curve does not fit.
For those we give the range of text sizes over which the share read correctly passes half.

\begin{table}[t]
  \centering
  \footnotesize
  \setlength{\tabcolsep}{3.2pt}
  \begin{tabular}{@{}llrrrrc@{}}
    \toprule
    & & \multicolumn{4}{c}{Canvas long side (px)} & \\
    \cmidrule(lr){3-6}
    Model & Setting & 1024 & 2048 & 4096 & 8000 & H1 \\
    \midrule
    Claude Haiku 5.5 & native & 4.7 & 4.8 & 8.1 & 14.4 & yes \\
     & standard & 4.4 & 6.8 & 12.5 & 24.8 & yes \\
    Claude Sonnet 5.5 & native & 4.2 & 4.4 & 6.5 & 12.0 & no \\
     & standard & 4.3 & 5.5 & 9.9 & 21.3 & no \\
    \midrule
    Gemini 3.5 Flash-Lite & high & 4.8 & 4.9 & 8.2 & 13.7 & no \\
     & ultra high & 4.2 & 4.4 & 5.4 & 10.8 & no \\
    Gemini 3.8 Flash & low & 4.4 & 7.2 & 13.2 & 26.7 & no \\
     & medium & 4.3 & 5.2 & 9.9 & 19.7 & no \\
     & high & 4.2 & 4.4 & 7.5 & 13.2 & no \\
     & ultra high & 4.2 & 4.2 & 5.2 & 10.8 & no \\
    \midrule
    GPT-6 Luna & auto & 4 to 6 & 4.2 & 4 to 6 & rejected & n/a \\
     & high & 4.1 & 5.0 & 10.2 & 17.9 & yes \\
    GPT-6.1 Sol & auto & 4.3 & 4.5 & 4 to 6 & rejected & n/a \\
     & high & 4.2 & 5.0 & 8.5 & 15.7 & no \\
    \midrule
    Qwen3.8-27B & default & 5.4 & 6.2 & 9.0 & 15.1 & no \\
    \bottomrule
  \end{tabular}
  \caption{\textbf{Readable cap height in pixels on each canvas.}
  Where the fit does not cross half within our sizes, the cell gives the sizes between which reads did.
  H1: whether tokens fit the reads better and the canvases agree in tokens (n/a: OpenAI shrinks nothing).}
  \label{tab:readable}
\end{table}

Is shrinking the whole story?
If it is, a letter that ends up covering the same number of tokens should be equally easy to read, whatever canvas it started on.
The four curves in \cref{fig:eye} should then line up when text size is counted in the tokens the model receives, instead of in original pixels.
This is our first hypothesis, H1, and for Claude and OpenAI the curves move much closer (\cref{fig:eye}, bottom).
We set two tests for each model and setting.
First, measuring text in tokens had to fit the reads better than measuring it in pixels, by at least 10 points of AIC, a standard criterion that weighs fit against model size~\cite{akaike1974new}.
Second, the readable sizes from the separate canvases, counted in tokens, had to agree within 25\%.
The token scale fit better wherever a service shrinks images, by 39 to 358 points.
Three settings passed both tests: Haiku at both settings and Luna at high.
Sonnet at both settings and Sol at high came close: across canvases, their largest readable size in tokens was 25\% to 40\% above their smallest.
Many single-canvas curves jump from few correct reads to nearly all correct between adjacent sizes, so these spreads depend somewhat on the fit's penalty: for any penalty from 0.0001 to 0.1, Haiku at standard and Luna at high pass and Sonnet fails, while Haiku at native and Sol at high fall on either side of the line.
Gemini failed the second test, for reasons we explain in \cref{sec:gemini}.
Qwen failed it too, and shrinking alone cannot explain why: its host keeps the three smaller canvases whole, yet its readable size grew from 5.4 to 9.0 pixels.

Claude's two settings give a second, direct test.
On our two largest canvases, the token limit, not the long-side limit, decides how far each setting shrinks the image, so the native setting keeps 1.74 times as many pixels per side.
If shrinking is the whole story, the readable size at the standard setting should be 1.74 times the readable size at the native setting.
On the 8000 canvas, this ratio was 1.73 for Haiku and 1.78 for Sonnet, close to the prediction.
On the 4096 canvas, both measured about 1.55, below the prediction; Sonnet's 95\% interval excludes it, and Haiku's only just reaches it.

\begin{figure}[t]
  \centering
  \includegraphics[width=\linewidth]{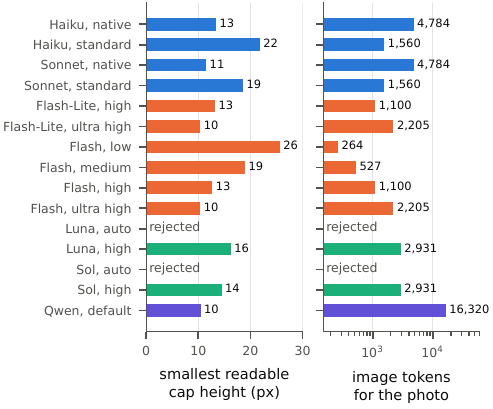}
  \caption{\textbf{What survives on one 8K photo.}
  Left: the predicted cap height each model reads half the time on a 7680 by 4320 photo.
  Right: the number of tokens the photo becomes.}
  \label{fig:eightk}
\end{figure}

\subsection{Gemini sees more than its token count}
\label{sec:gemini}

Gemini's curves did not line up in tokens, except roughly at its lowest setting.
At high, Flash could read letters 0.15 of a token tall on the 1024 canvas, but letters only 0.06 of a token tall on the 8000 canvas.
In other words, a token on a large image carries more detail than a token on a small one, and the token count understates what Gemini sees.

A simpler picture fits the data, at least for Flash.
Flash behaves as if it shrinks each image to a fixed long side, set by the resolution setting, and leaves smaller images alone.
To estimate this long side, we assume that Gemini keeps the 1024 canvas whole at ultra high.
The readable size there is then what Flash needs in the pixels it receives, and how much larger the readable size is on another canvas shows how far that canvas was shrunk.
On the two largest canvases, the long side Flash keeps comes out at about 1,300 pixels at low and 1,700 at medium.
It is about 2,300 to 2,600 at high and 3,100 to 3,400 at ultra high.
Spread over the token grid, each token therefore covers a patch roughly 60 to 70 pixels square.
Flash-Lite gives similar values on the large canvases.
On the two smaller canvases, though, its readable size is 4.8 and 4.9 pixels at high, against 4.2 and 4.4 at ultra high, which the simpler picture does not predict.
The 4.2-pixel size we compare against sits near the 4-pixel floor of our scale, so these estimated long sides may be too large.

\subsection{What a whole 8K photo keeps}
\label{sec:eightk}

The same fits tell an agent how small text can be on a single 8K photo (\cref{fig:eightk}).
For most models we use, for each setting, the single curve fit in tokens across all four canvases, like those dashed in \cref{fig:eye}.
For Gemini we use the fixed long side of \cref{sec:gemini}.
Under that long side, a 7680 by 4320 photo is shrunk by almost the same factor as our 8000 canvas, so we take the size measured on that canvas and scale it to the photo's width.
Each size below is the cap height read half the time, and the fits predict that smaller text is read less often.
On a 7680 by 4320 photo, Sonnet would need letters 11 pixels tall and Haiku 13.
Flash and Flash-Lite at high, which matches the default, would need 13 pixels, and Flash would need 26 at its lowest setting.
OpenAI's default rejects such a photo, and its high setting would need 14 to 16 pixels.
Qwen's fit gives 10 pixels, but this estimate is loose, since its readable size grew even on canvases its host keeps whole.

\subsection{Models rarely claim to read erased text}
\label{sec:claims}

We also asked whether models pretend to read text they cannot see.
On printed codes, they mostly do not.
Our second hypothesis, H2, expected models to answer confidently when the text was far too small to read.
We took every image whose text, in received tokens, was under half the readable size from the single curve for its model and setting, and counted the answer as confident if the model called the code readable or gave a confidence of 70 or more.
H2 held if the 95\% interval of this share lay above 70\% and the test, corrected for multiple testing, passed.
It held for Sonnet and Sol, at 98\% each, and failed for the other five models, at 0\% to 14\%.
But the two passes are misleading.
In every such confident answer, the model said the code was \emph{not} readable, and its high confidence was in that judgment.
Counting only claims to have read the code, H2 fails for all seven models.
No model read a single code correctly in this range.

\section{Real photos and the zoom check}
\label{sec:photos}

\begin{figure*}[t]
  \centering
  \includegraphics[width=\linewidth]{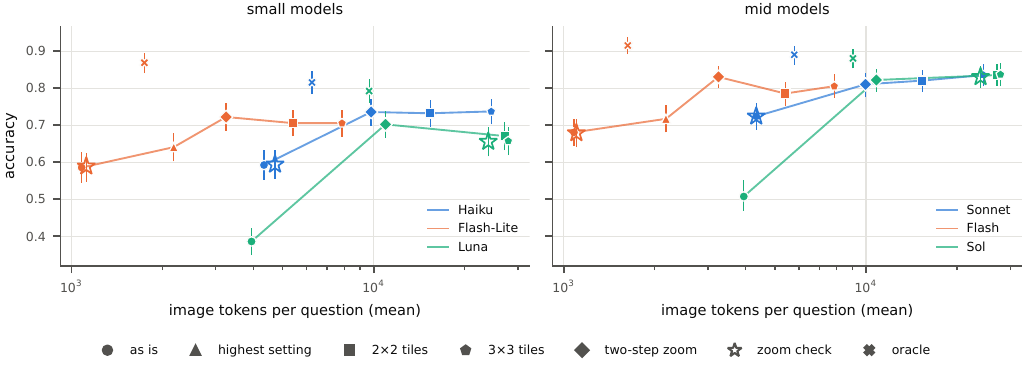}
  \caption{\textbf{Tiling and zooming trade image tokens for accuracy.}
  Tokens follow each service's rules, count nothing for a rejected call and include both zoom calls, so compare them only within a model.
  As in \cref{tab:accuracy}, 3$\times$3 falls back to 2$\times$2 below 4000 pixels.
  The optimistic oracle takes the correct way with the fewest tokens, or the cheapest way if none is correct, and bars are 95\% bootstrap intervals.}
  \label{fig:accuracy}
\end{figure*}

Printed codes are idealized.
On real photos, the object sits somewhere in a whole scene, and the model can guess.

\begin{table}[t]
  \centering
  \footnotesize
  \setlength{\tabcolsep}{2.6pt}
  \begin{tabular}{@{}lrrrrrr@{}}
    \toprule
    Model & As is & Highest & 2$\times$2 & 3$\times$3 & Zoom & Check \\
    \midrule
    Claude Haiku 5.5 & 59.2 & n/a & 73.2 & \textbf{73.7} & 73.5 & 59.4 \\
    Gemini 3.5 Flash-Lite & 58.6 & 64.1 & 70.5 & 70.5 & \textbf{72.2} & 58.9 \\
    GPT-6 Luna & 38.6 & n/a & 67.1 & 65.7 & \textbf{70.2} & 65.6 \\
    \midrule
    Claude Sonnet 5.5 & 72.4 & n/a & 82.0 & \textbf{83.5} & 81.0 & 72.4 \\
    Gemini 3.8 Flash & 68.1 & 71.7 & 78.5 & 80.5 & \textbf{83.0} & 67.9 \\
    GPT-6.1 Sol & 50.7 & n/a & 83.5 & \textbf{83.7} & 82.2 & 83.0 \\
    \midrule
    Qwen3.8-27B & \textbf{71.5} & n/a & n/a & n/a & n/a & n/a \\
    \bottomrule
  \end{tabular}
  \caption{\textbf{Accuracy (\%) on 601 questions.}
  Highest is Gemini's ultra high setting, and 3$\times$3 falls back to 2$\times$2 for photos under 4000 pixels.
  Check is the zoom check's pick, and bold is best per model.}
  \label{tab:accuracy}
\end{table}

\subsection{Tiling and zooming raise accuracy}
\label{sec:tiling}

Tiling recovers much of the lost detail.
Sent as is, the photos gave accuracies of 38.6\% to 72.4\% (\cref{tab:accuracy}).
Sending 3 by 3 tiles, or 2 by 2 tiles for photos under 4000 pixels, raised accuracy by 14.5 points for Haiku, 11.1 for Sonnet, 12.0 for Flash-Lite and 12.5 for Flash.
For these models the gain was positive on every question set: 8 to 18 points on V*Bench, 10 to 18 on HR-Bench, 14 to 21 on remote sensing and 2 to 11 on monitoring.
On remote sensing, 3 by 3 tiles lifted Sonnet from 60.0\% to 80.8\%.
The highest setting added only 5.5 points for Flash-Lite and 3.7 for Flash.

Not all of this gain comes from undoing the service's shrinking.
Claude receives the photo at full size for 304 of the 601 questions, and even there tiles added 12.2 points for Haiku and 5.3 for Sonnet.
On the other 297 questions, whose photo Claude shrinks, tiles added 16.8 and 17.2 points.
We believe tiles also help because a small object fills more of each image the model sees.
Earlier work also found that adding an enlarged crop of a small object helps~\cite{zhang2025mllms}.

OpenAI's default rejected the photo for 215 of the 601 questions, including 144 of the 150 HR-Bench questions.
On the 386 photos it accepted, Sol scored 79.0\% as is, so its low overall score reflects rejections more than poor perception.
Tiles avoid the rejection, and they lifted Sol from 50.7\% to 83.7\% overall.
For comparison, Qwen, whose host keeps far more pixels than Claude or Gemini, scored 71.5\% as is.

The two-step zoom needs far fewer image tokens than tiles.
Counting both calls, it used about 40\% of the image tokens of 3 by 3 tiles (\cref{fig:accuracy}).
On Gemini it scored highest of all five ways: 72.2\% for Flash-Lite and 83.0\% for Flash, against 70.5\% and 80.5\% for 3 by 3 tiles, though the paired 95\% intervals of these gains include zero.
It did this even though Gemini did not always return a valid box: four fractions between 0 and 1, with left less than right and top less than bottom.
Flash-Lite did so for 87\% of the questions and Flash for 96\%, and for the rest the second call sent the whole photo.
The other four models returned a valid box every time.
On Claude, the zoom nearly matched 3 by 3 tiles with Haiku, at 73.5\% against 73.7\%, but trailed them with Sonnet, at 81.0\% against 83.5\%.
Luna did best with the zoom, 4.5 points above 3 by 3 tiles with a paired 95\% interval of 1.5 to 7.5 points, and Sol came within 1.5 points of its best tiles.
On remote sensing, the two-step zoom scored highest for five of the six models, lifting Flash from 61.5\% to 79.2\% and Sol from 36.2\% to 84.6\%.
OpenAI's default rejected the zoom's second call for one HR-Bench question with Luna and three with Sol, because the crop, with its margin, covered the whole photo, which the default rejects.

\begin{figure}[t]
  \centering
  \includegraphics[width=\linewidth]{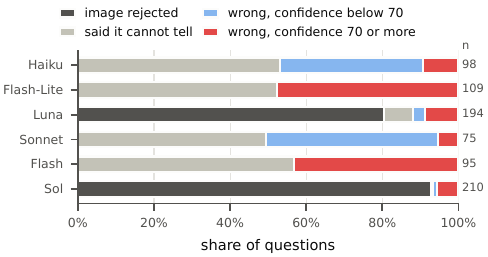}
  \caption{\textbf{What models answer when the detail is lost.}
  Answers to the photo sent as is, on questions that tiles got right and the photo as is got wrong.
  The number of such questions is on the right.}
  \label{fig:lost}
\end{figure}

\subsection{When the detail is lost, Gemini stays confident}
\label{sec:lost}

What does a model say when the detail it needs is lost?
\Cref{fig:lost} looks at the questions where tiles found the right answer and the full photo did not.
Tiles cut from the same photo found the answer, so the detail was there, and the full photo most likely lost it.
Chance explains some of these cases, since answers also flipped the other way, right as is and wrong with tiles, on 8 to 37 questions per model.
For OpenAI's models, most of these photos had been rejected.
Claude and Gemini both said they could not tell in about half of these cases: 53\% for Haiku, 49\% for Sonnet, 52\% for Flash-Lite and 57\% for Flash.
The difference lies in their other wrong answers.
Claude gave most of them with a confidence below 70, so only 9\% of these cases for Haiku and 5\% for Sonnet were confident wrong answers.
Gemini gave every one of them with a confidence of 70 or more, which made 48\% of these cases confident wrong answers for Flash-Lite and 43\% for Flash.
With our prompt and settings, Claude's stated confidence warned of a lost detail, and Gemini's did not.

Gemini's confidence here differs from its answers on printed codes too small to read, where it claimed a read or high confidence at most 14\% of the time (\cref{sec:claims}).
We believe the format explains much of the difference: a code must be read exactly, while a multiple-choice question always offers a plausible guess.

\subsection{A check at 16 pixels misses most of the gain}
\label{sec:check}

Could an agent have known before the call which photos needed tiles?
Our third hypothesis, H3, was that the zoom check would come within 2 points of 3 by 3 tiles, with 95\% confidence, while using at most 60\% of their tokens.
It failed for all six models that received tiles.
For Claude and Gemini, the check judged a 16-pixel detail readable on almost every photo.
That follows from the eye exam: at default settings, the readable sizes the check uses for an 8K photo are at most about 14 pixels, close to those in \cref{fig:eightk}.
So the check sent the photo as is for at least 96\% of the questions, and scored 11 to 14 points below tiles.
For OpenAI, it tiled the rejected photos and came within 1 point of tiles on average, but its 95\% interval reached 2.7 to 3.5 points below them, and it used 86\% of their tokens.
Our fourth hypothesis, H4, was that the photos the check flags, those it would not send as is, score lower when sent as is.
It also failed.
For Claude and Gemini, the check flagged at most 21 of the 601 questions, too few to test well, and for Flash the flagged ones even scored higher.
For OpenAI, it flagged exactly the rejected questions, which the test omits, and for Qwen none.

V*Bench, the only benchmark that marks where each target object is, shows one reason the check failed.
The median target there is 36 pixels across its shorter side, well above the check's 16 pixels, yet tiles still raised Claude and Gemini accuracy there by 8 to 18 points.
We believe the main reason is the check's bar.
The readable size marks where printed text is read half the time, and a detail that size is far from enough to answer a question about an object.

Checking for smaller details helps.
At 8 pixels, the check recovered about half of the gain from tiles for both Claude models, using about half the tokens.
For Gemini it recovered about half for Flash-Lite and a quarter for Flash.
Even so, it did not beat the two-step zoom on Claude.
Two simpler signals need no detail size at all.
The first asks only whether the service will shrink the photo, using \cref{tab:rules}.
Tiling just the 297 photos Claude shrinks would add 8.3 points for Haiku and 8.5 for Sonnet, over half of the full gain, but would still use over 90\% of the image tokens of tiling every photo.
The second signal is the reply itself.
On the lost-detail questions of \cref{fig:lost}, Claude said it could not tell, or gave a confidence under 70, in 91\% of the cases for Haiku and 95\% for Sonnet, against only 52\% for Flash-Lite and 57\% for Flash.
An agent that sends the photo as is and tiles it only after such a reply reaches 72.4\% with Haiku and 82.9\% with Sonnet, within 1.4 points of always tiling, with 62\% to 70\% of the image tokens.
On Gemini, it falls 2.7 to 3.7 points behind always tiling, as its replies flag fewer lost details.
With the two-step zoom as the fallback instead, Claude stays within 3 points of always tiling with 36\% to 39\% of those tokens.

\section{Related work}
\label{sec:related}

\paragraph{Zooming into large images.}
Several methods find small details by searching or cropping.
SEAL, from the V* paper, searches an image under the guidance of a language model~\cite{wu2024vstar}, ZoomEye explores a tree of zoomed views with open models~\cite{shen2025zoomeye}, and DC\textsuperscript{2} splits 4K and 8K images into patches and combines what the model says about each~\cite{wang2025divide}.
Zoomer chooses which parts of an image to keep under a limit on image tokens, and works with closed models~\cite{qian2025zoomer}.
DeepEyes learns through reinforcement learning to crop the regions it needs while it reasons~\cite{zheng2026deepeyes}.
Other methods crop using a model's attention and gradient maps, which online services do not expose~\cite{zhang2025mllms}.
Closest to our question, one study works out from vendors' documented rules how many image tokens an object spans, links missed small objects to that number, and gives a rule for when to enlarge or split an image~\cite{shi2026why}.
It also proves, under stated assumptions, when splitting an image into views cannot lower recall.
Benchmarks such as V*Bench, HR-Bench and MME-RealWorld collect questions about small details in large images~\cite{wu2024vstar,wang2025divide,zhang2025mmerealworld}, and RSHR-Bench does the same for remote sensing~\cite{dang2025benchmark}.
None of them measures what each service's own resizing and rejections do to the smallest readable detail and to answers about real photos.

\paragraph{Resolution and preprocessing.}
FineSightBench finds the smallest patterns models perceive, but on 448 by 448 canvases that no service shrinks~\cite{li2026lastvisiblepixel}.
Res-Bench tests whether accuracy stays stable across input resolutions~\cite{li2026resbench}.
Upscaling or padding an image without changing its content can lower the accuracy of open models~\cite{schild2026waldo}.
Small resizes that change how an image is tiled can also turn right answers into wrong ones~\cite{lin2026readingright}.
Other work picks the input resolution for each task~\cite{luo2025taskaware}, or relates accuracy to the size of the language model and the number of image tokens~\cite{zhao2026notallerror}.
In document question answering, giving retrieved pages more image tokens raises accuracy~\cite{jo2026loudclear}.
One study compares token counts for code shown as images across providers, without measuring accuracy~\cite{bhalgami2026pixels}, and another shows that attackers can exploit image scaling in production systems~\cite{morozova2025weaponizing}.
We instead measure, for the services and settings we tested, which detail sizes reach the model.

\paragraph{Answers without the evidence.}
Models describe images they never received~\cite{asadi2026mirage} and give diagnoses without the image~\cite{vohra2026absentbyte}.
In medical imaging, they stay confident as image quality drops~\cite{liu2026medqdeg}.
In our setting the image is sent in full, yet part of it never reaches the model, and models differ in how often they then answer wrongly with confidence.

\section{Practical rules}
\label{sec:rules_of_thumb}

Our results suggest five rules for an agent.
\begin{enumerate}
  \item On Claude, send the photo as is and tile it whenever the reply says it cannot tell or gives a confidence under 70. This came within 1.4 points of tiling every photo with 62\% to 70\% of the tokens. Where accuracy matters most, tile every photo.
  \item On Gemini, always use the two-step zoom, since its confident replies hide lost details. It matched or beat tiles with about 40\% of their tokens and beat the highest setting. Check that each box is valid (\cref{sec:tiling}), and send the whole photo if it is not.
  \item On OpenAI's default setting, use tiles for any image above 30,000 patches, since the service rejects it, or a zoom whose crop stays under that limit.
  \item Before the call, use \cref{tab:rules} to see whether the service will shrink the image. A size check like ours should use a detail size well below 16 pixels.
  \item Except on Claude, whose confidence warned of lost details in our tests, treat a confident answer about a small detail with suspicion.
\end{enumerate}

\section{Limitations}
\label{sec:limits}

Vendors can change their rules, but the released eye exam can measure a service again.
We collected one answer per question and setting, so our intervals cover the choice of questions, not the variation between calls.
All models ran at a low reasoning level, and stated confidence may not mean the same thing across vendors.
Qwen received photos only as is, so we cannot say how much tiles help a model that keeps far more pixels.
We tested no medical or radar images.

\section{Broader impact}
\label{sec:impact}

Vision-language services are asked to read satellite, security and medical images, and our results show that a confident answer may rest on detail the model never saw.
The simple rules above should make such errors less likely.

\section{Conclusion}
\label{sec:conclusion}

Online services shrink, squeeze or reject large images before the model sees them.
On an 8K photo at default settings, our fits predict that text under 10 to 13 pixels is read less than half the time.
On real questions about small objects, up to nine tiles add 11.1 to 14.5 percentage points for Claude and Gemini, and the two-step zoom comes within 2.5 points of tiles or beats them with about 40\% of their image tokens.
When the full photo misses a detail, Gemini gives a confident wrong answer in 43\% to 48\% of cases, against 5\% to 9\% for Claude, so on Claude the reply itself tells an agent when to zoom.
Elsewhere, since tiles helped even on V*Bench, whose median target is 36 pixels across its shorter side, we would zoom in for small details in large photos.

\paragraph{Acknowledgments.}
This work was supported in part by U.S. NIH grant R35GM158094.
The Gemini calls in this study were supported by a Gemini Academic Program award of Google Cloud credits from Google.
Google had no role in study design, execution, analysis or reporting, and did not review or influence this work.
Generative AI tools assisted in preparing this work, and the authors reviewed all content and take full responsibility for it.

{
  \small
  \bibliographystyle{ieeenat_fullname}
  \bibliography{refs}
}

\end{document}